\documentclass[10pt,twocolumn,letterpaper]{article}

\usepackage[pagenumbers]{cvpr}
\usepackage{times}
\usepackage{epsfig}
\usepackage{graphicx}
\usepackage{amsmath,amssymb}
\usepackage{booktabs}
\usepackage{multirow}
\usepackage{array}
\usepackage{microtype}
\usepackage{enumitem}
\usepackage{pifont}
\usepackage{url}
\usepackage{mathtools}
\usepackage{float}
\usepackage{booktabs}
\usepackage{makecell}
\usepackage{graphicx}
\usepackage{tabularx}
\usepackage{adjustbox}

\DeclareCaptionLabelSeparator{fixedperiod}{.\nobreak\hspace{0.25em}}

\definecolor{cvprblue}{rgb}{0.21,0.49,0.74}
\definecolor{diagred}{RGB}{168,44,44}
\definecolor{directgreen}{RGB}{33,119,82}
\definecolor{draftorange}{RGB}{190,96,15}

\usepackage{xr-hyper}
\usepackage[pagebackref,breaklinks,colorlinks,allcolors=cvprblue]{hyperref}

\def\confName{CVPR}
\def\confYear{2027}

\newcommand{\concor}{\texttt{\textbf{CONCOR}}}

\title{CONCOR: Benchmarking Contrastive Decoding for Visual Hallucination Mitigation Beyond Final-Output Performance}

\author{Ting Chen$^{1,*}$
\qquad
Geng Li$^{2,*}$
\qquad
Guohao Chen$^{2}$
\qquad
Hanjia Lyu$^{1}$\\
{\small $^{1}$Singapore Management University}
\qquad
{\small $^{2}$Nanyang Technological University}\\
{\small $^{*}$Equal contribution}
}

\begin{document}

\ifdefined\SUPPLEMENTARY
    \maketitlesupplementary
\else
    \maketitle

\begin{abstract}
Contrastive decoding (CD) mitigates object hallucination in multimodal large language models (MLLMs) by contrasting token distributions under different conditions to adjust token scores. 
However, existing generative and discriminative benchmarks primarily measure final task scores, which can be confounded by candidate filtering, changes in generation trajectories, and response biases,  making it difficult to isolate the effects of the underlying contrastive adjustments. 
To address these limitations while complementing existing evaluations, we introduce \concor{} (\underline{Con}trastive \underline{Cor}rection Evaluation), a controlled benchmark that directly evaluates the competition between ground-truth and different types of hallucinated candidates using the Contrastive Grounding Margin (CGM). 
CGM also enables analysis of directional consistency, composition, and reliability of contrastive corrections.
Our evaluation reveals that existing CD methods exhibit distinct correction behaviors: some consistently favor ground-truth candidates, whereas others produce strong corrections but also introduce more errors. Crucially, these differences help explain their contrasting responses to increased contrast strength, which improves some methods while degrading others.
These findings highlight the importance of evaluating the contrastive correction itself, rather than relying solely on final-output performance, and \concor{} provides a controlled and interpretable framework for understanding and improving CD methods for hallucination mitigation.
\end{abstract}

\begin{figure}[!t]
\centering
\includegraphics[width=\columnwidth]{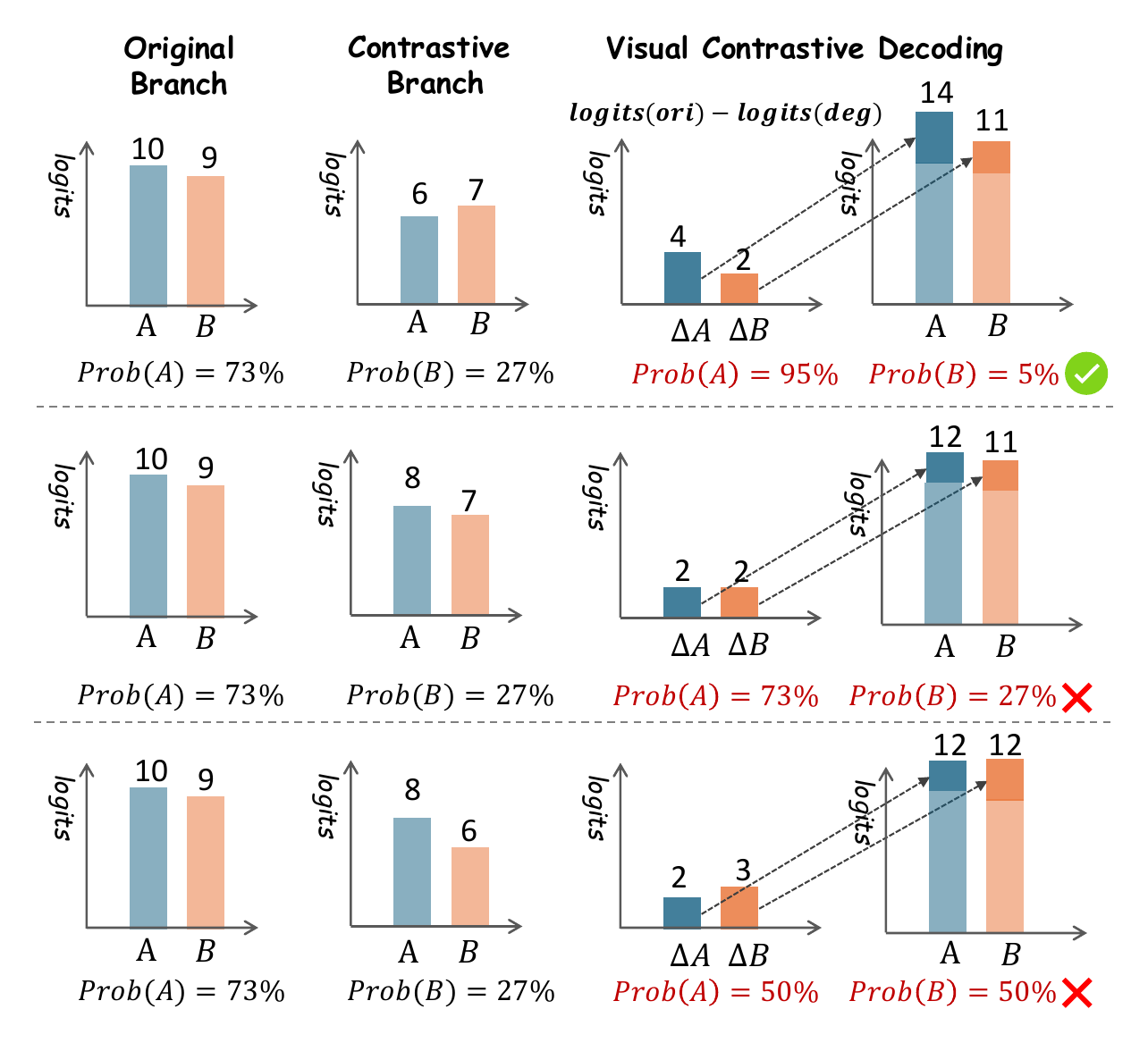}
\caption{Three possible outcomes of visual contrastive decoding. 
CD increases the relative preference for a visually grounded token $A$ over a hallucinated competitor $B$ only when its original-to-contrastive logit gap is larger, \textit{i.e.,} $\Delta_A>\Delta_B$.
When $\Delta_A=\Delta_B$, their relative preference remains unchanged; when $\Delta_A<\Delta_B$, CD instead favors $B$ relative to $A$.
Probabilities are computed by applying softmax over the two illustrated candidates, with $\alpha=1$.}
\label{fig:cd_outcomes}
\end{figure}

\section{Introduction}
\label{sec:intro}

Recent advances in integrating vision encoders with large language models have endowed multimodal large language models (MLLMs) with broad, general-purpose capabilities in image and video description~\cite{maaz2024video}, visual question answering~\cite{dai2023instructblip}, multimodal dialogue~\cite{Liu_2024_CVPR}, visual grounding~\cite{peng2024grounding}, document understanding~\cite{hu2024mplug}, and visual reasoning~\cite{shao2024visual,icpr2024}. MLLMs are consequently emerging as an important foundation for connecting visual perception with natural-language interaction~\cite{li2023blip}. Nevertheless, when visual evidence is weak or underutilized, these models often over-rely on linguistic priors and scene co-occurrence patterns, producing plausible yet visually unsupported objects, attributes, or relations~\cite{guan2024hallusionbench,yang2025understanding,jiang2025devils,lee-etal-2025-vlind,tist2025}. This phenomenon, commonly known as visual hallucination~\cite{jing2024faithscore,min-etal-2025-mitigating}, undermines the trustworthiness of model outputs and limits deployment in high-reliability settings such as medical assistance and autonomous driving ~\cite{chen2024advanced,xie2025vlms,zhu2025can}.

Researchers have explored training-time alignment~\cite{jiang2024hallucination,sun2024aligning,yu2024rlhf,yu2025rlaif,yoon2025stop}, post-hoc correction~\cite{lee2024volcano,yin2024woodpecker,zhou2024analyzing}, and decoding-time intervention to mitigate visual hallucinations~\cite{leng2024mitigating,zou2024look,huang2024opera,wan2025only}. Among these approaches, contrastive decoding (CD) has attracted considerable attention because it requires no additional training and operates directly at inference time~\cite{leng2024mitigating,suo2025octopus}. CD contrasts token distributions produced under different conditions and uses their differences to adjust next-token scores~\cite{kim2025fuzzy}. Mechanistically, CD provides effective correction only when the contrastive adjustment increases the relative preference  for a visually grounded candidate over a candidate that lacks visual support. If this relative preference instead decreases, the contrastive intervention may reinforce the erroneous candidate.

Existing methods typically assess CD's ability to mitigate hallucinations through scores on generative and discriminative tasks~\cite{leng2024mitigating,wang2024mitigating,favero2024multi}. However, these scores are influenced by both decoding choices and task design.
First, CD can benefit from a shortcut to higher benchmark scores by adopting a decoding strategy that filters out low-probability hallucinated tokens~\cite{yin2026mirage}. Second, generative evaluations apply CD throughout generation, where revisions to non-target content can alter the subsequent generation context and thereby affect later object decisions~\cite{wang2023amber,zhong2024investigating,zheng2025lvlms}.
Third, accuracy in discriminative evaluations is sensitive to Yes/No response tendencies~\cite{li2023evaluating}. CD may therefore improve accuracy by increasing the rate of affirmative responses, allowing changes in response tendency to be counted as gains in hallucination mitigation. Together, these factors make it difficult to isolate the direct effect of contrastive intervention on object decisions from the reported final task scores.

To address these limitations while complementing existing evaluations, we introduce \concor{} (\underline{Con}trastive \underline{Cor}rection Evaluation), a controlled benchmark for directly evaluating the competition between ground-truth and hallucinated candidates; see Appendix~\ref{appendix:benchmark_details} for details.
\concor{} constructs candidate sets containing ground-truth objects and different types of hallucinated objects selected according to their co-occurrence relationships with the image content.
All candidates are evaluated at the same decoding position and context, using logits extracted from the original and contrastive branches before candidate filtering. This design isolates the contrastive adjustment from downstream generation trajectories, candidate filtering, and response tendencies.
Within \concor, we introduce the Contrastive Grounding Margin (CGM) to diagnose whether and how CD strengthens the relative advantage of ground-truth over hallucinated candidates.
Beyond measuring the direction and magnitude of contrastive correction, CGM provides three complementary diagnostic properties including directional consistency, correction composition, and correction reliability.
Our contributions are threefold:
\begin{itemize}
    \item We identify how decoding constraints, generation trajectories, and response bias can confound existing generative and discriminative evaluations, obscuring the underlying contrastive correction produced by CD.

    \item We introduce \concor{}, a controlled benchmark for directly evaluating competition between ground-truth and different types of hallucinated candidates, together with CGM for measuring and diagnosing contrastive correction.

    \item We reveal substantial differences in the directional consistency, composition, and reliability of existing CD methods and how that these differences help explain why increasing contrast strength benefits some methods while degrading others, with the main patterns remaining robust across models and task settings.
\end{itemize}

\section{Related Work}
\label{sec:related}

\paragraph{Contrastive decoding.}
Existing contrastive decoding methods differ mainly in how they obtain the distributions used for comparison. VCD~\cite{leng2024mitigating} perturbs the input image,
ICD~\cite{wang2024mitigating} introduces disturbance instructions , while SID~\cite{huo2025self} and AvisC~\cite{woo2025don} manipulate visual representations based on token relevance or attention. 
Other approaches construct contrast through confidence-aware logit calibration~\cite{kim2025fuzzy}, image-unconditioned predictions~\cite{favero2024multi}, or intermediate-layer representations~\cite{chen2026yard}.
Despite these differences, these methods share the principle of contrasting alternative output distributions to adjust token predictions.

\vspace{-3mm}
\paragraph{Hallucination evaluation.}
Existing benchmarks evaluate multimodal hallucination through generative, discriminative, and combined settings. CHAIR measures object hallucination in image captioning by comparing generated object mentions with ground-truth image annotations~\cite{rohrbach2018object}. It reports both the proportion of captions containing hallucinated objects and the proportion of hallucinated objects among all object mentions. POPE formulates object hallucination evaluation as binary existence questions and constructs negative samples using random, popular, and adversarial sampling strategies~\cite{li2023evaluating}. AMBER provides a broader benchmark covering both generative and discriminative hallucination without relying on an external LLM evaluator. Its evaluation includes hallucinations involving objects, attributes, and relations~\cite{wang2023amber}. Together, these benchmarks provide complementary measurements of object-level and response-level visual faithfulness and are widely used to assess hallucination mitigation methods in MLLMs~\cite{fazli2026caac}.

\section{Preliminaries}
\label{sec:preliminaries}

Given an instruction \(x\), an image \(v\), and a generation prefix \(y_{<t}\), let \(\ell_o(w \mid y_{<t}, x, v)\) denote the logit assigned to candidate token \(w\) by the original branch, and let \(\ell_d(w \mid y_{<t}, x, v')\) denote the corresponding logit produced by the contrastive branch with input \(v'\). Visual contrastive decoding computes
\begin{equation}
\ell_{\mathrm{CD}}(w)=(1+\alpha)\ell_o(w)-\alpha\ell_d(w)=\ell_o(w)+\alpha\Delta_w,
\label{eq:cd}
\end{equation}
where \(\Delta_w=\ell_o(w)-\ell_d(w)\), and \(\alpha>0\) controls the contrast strength. Thus, \(\alpha\Delta_w\) is the additive adjustment that CD applies to the original logit of token \(w\).

To examine whether this adjustment favors a visually grounded token, consider two candidates competing at the same decoding step under the same prefix. Let \(A\) denote a factual token supported by the visual content, and let \(B\) denote a hallucinated competitor that lacks visual support but may still receive a high score due to the model's language prior. Define their original-to-contrastive logit gaps as \(\Delta_A=\ell_o(A)-\ell_d(A)\) and \(\Delta_B=\ell_o(B)-\ell_d(B)\), respectively. The change in their pairwise logit margin induced by CD is
\begin{equation}
\left[\ell_{\mathrm{CD}}(A)-\ell_{\mathrm{CD}}(B)\right]-\left[\ell_o(A)-\ell_o(B)\right]=\alpha(\Delta_A-\Delta_B).
\label{eq:pairwise_margin}
\end{equation}
Because the pairwise probability ratio under softmax is determined by the corresponding logit difference, Eq.~\eqref{eq:pairwise_margin} shows that, as illustrated in Figure~\ref{fig:cd_outcomes}, CD increases the relative preference for \(A\) over \(B\) only when \(\Delta_A>\Delta_B\); otherwise, the relative preference remains unchanged or decreases.

\begin{figure}[t]
\centering
\includegraphics[width=\columnwidth]{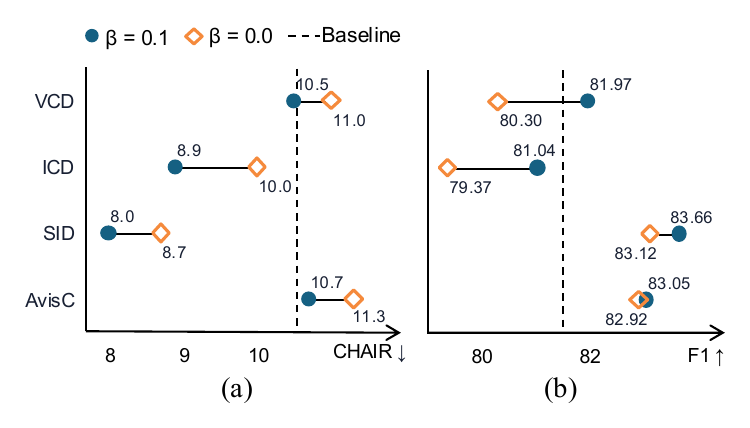}
\caption{
Effects of Adaptive Plausibility Constraint (APC) on (a) AMBER generation and (b) POPE discrimination across four CD methods.
Blue circles indicate APC enabled, while orange diamonds indicate APC disabled. 
Dashed lines represent baseline performance without CD. }
\label{fig:apc}
\end{figure}

\section{Limitations of Existing CD Evaluations}
\label{sec:rethinking}

Effective CD requires the contrastive branch to provide a corrective signal favoring visually grounded over hallucinated object candidates. However, existing evaluations primarily assess CD using downstream generative and discriminative task scores, which do not necessarily isolate the effectiveness of this signal.

\textbf{\textit{Observation 1: Decoding constraints confound the evaluation of CD's corrective ability.}}
Adaptive Plausibility Constraint (APC), commonly used in contrastive decoding methods~\cite{leng2024mitigating,wang2024mitigating,huo2025self}, restricts the candidate tokens using the original branch's predicted probabilities.
By filtering out low-probability candidates before final selection, APC can affect hallucination-related performance without necessarily strengthening the contrastive branch's corrective signal.
As shown in Figure~\ref{fig:apc}, enabling APC consistently reduces CHAIR and improves F1 across four visual contrastive decoding methods, while leaving their contrastive branch constructions unchanged. Experimental details are provided in Appendix~\ref{appendix: amber and pope}.
 These results demonstrate that downstream improvements can arise from decoding constraints rather than solely from contrastive reweighting.
 Consequently, existing evaluations with APC enabled cannot isolate the corrective ability of the contrastive branch.

\begin{figure}[t]
\centering
\includegraphics[width=\columnwidth]{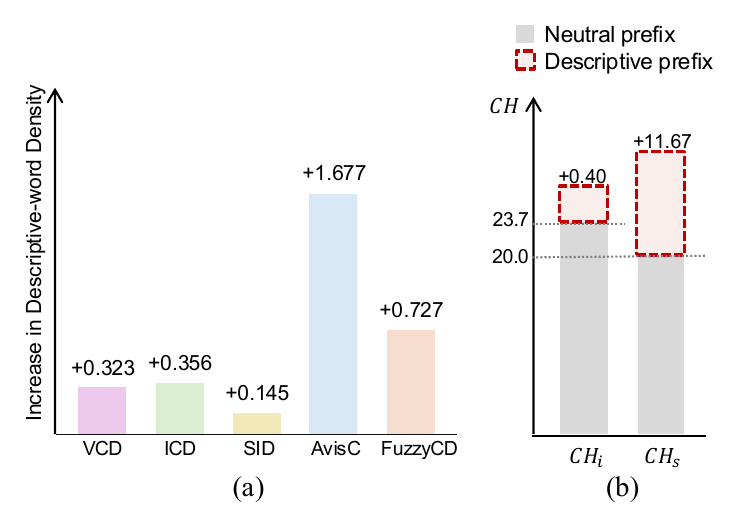}
\caption{
Effects of non-target content on generative evaluation.
\textbf{(a)} Change in descriptive-word density relative to the Baseline across five CD methods under sampling decoding. Values represent the mean increase in descriptive words per 100 generated words across 500 images. Descriptive words include adjectives, adverbs, numerals, and spatial-relation expressions. \textbf{(b)} Hallucination rates for continuations generated from neutral and descriptive prefixes, with continuation length matched. 
}
\label{fig:prefix-text-feature}
\end{figure}

\textbf{\textit{Observation 2: Generative evaluations confound object-level correction with the changes in non-target content.}}
Because CD is applied throughout autoregressive generation, it can alter not only object tokens but also non-target content, such as attributes and actions.
These changes become part of the generation prefix and may subsequently influence object predictions.
Figure~\ref{fig:prefix-text-feature}(a) shows that all five CD methods increase descriptive-word density relative to the Baseline under sampling decoding.
To examine whether such changes can affect subsequent hallucination, we compare continuations generated from a neutral prefix and a descriptive prefix that contains additional descriptive attributes but no additional object mentions. Experimental details are provided in Appendix~\ref{appendix: Non-target Descriptions on Generative}.
As shown in Figure~\ref{fig:prefix-text-feature}(b), changing from the neutral to the descriptive prefix increases $\mathrm{CHAIR}_s$ from $20.00\%$ to $31.67\%$ under matched continuation lengths, while $\mathrm{CHAIR}_i$ changes only slightly from $23.70\%$ to $24.10\%$.
This controlled intervention demonstrates that non-target prefix content can alter subsequent object hallucination. 
As a result, generative evaluations may conflate direct contrastive effect at object-token positions with indirect effects arising from changes in the generation trajectory, making it difficult to isolate the contrastive branch's object-level corrective ability.

\textbf{\textit{Observation 3: Discriminative evaluations can conflate response bias with improved hallucination mitigation.}}
Effective CD should improve discrimination between objects that are present and absent in the image: predictions of present objects should increase, while false predictions of absent objects should decrease. However, prior work~\cite{yin2026mirage} demonstrates that discriminative scores can also be improved through an affirmative-answer shortcut using Prompt-Based Adjustment (PBA) and Output Layer Modification (OLM).
Neither method constructs a contrastive branch; both achieve POPE gains comparable to CD by inducing a stronger tendency to answer \emph{Yes}, which can increase the aggregate score without necessarily improving discrimination between present and absent objects. 
Our results in Table~\ref{tab:yes_no_accuracy} further illustrate the difference between these two patterns. AvisC increases TPR, relative to the Baseline, from $74.36\%$ to $76.02\%$ while reducing FPR from $8.11\%$ to $7.04\%$. It therefore improves accuracy on both positive (Yes) and negative (No) questions. 
In contrast, SID and VCD increase both TPR and FPR, accompanied by increases in their overall \emph{Yes} rates from $41.23\%$ to $43.61\%$ and $43.50\%$, respectively. Their gains on positive questions therefore come at the cost of worse performance on negative questions, indicating a shift toward \emph{Yes} rather than uniformly improved discrimination. 
Such a shift can still yield higher aggregate scores: SID achieves $84.70\%$ overall accuracy, exceeding AvisC's $84.49\%$. 
Thus, aggregate discriminative scores can favor a method exhibiting a stronger affirmative-answer bias over one that better distinguishes present from absent objects, making them unreliable for isolating the hallucination-mitigation ability of CD.

\begin{table}[t]
\centering
\caption{Discriminative evaluation misattributes the affirmative-answer shortcut to gains in hallucination mitigation. AvisC improves TPR while reducing FPR, indicating genuine hallucination suppression. In contrast, SID and VCD increase both TPR and FPR, achieving higher overall scores through a Yes bias. Such shifts improve discriminative performance without necessarily reflecting better visual grounding.
See Appendix~\ref{appendix: amber and pope} for experimental details.}
\label{tab:yes_no_accuracy}
\scriptsize
\setlength{\tabcolsep}{2.5pt}
\renewcommand{\arraystretch}{1.05}

\begin{tabular}{lcccccc}
\toprule
Method
& TPR$\uparrow$
& FPR$\downarrow$
& Overall YR
& Yes Acc.$\uparrow$
& No Acc.$\uparrow$
& Overall Acc.$\uparrow$ \\
\midrule
Baseline & 74.36\% & 8.11\%  & 41.23\% & 74.36\% & 91.89\% & 83.12\% \\
SID      & 78.31\% & 8.91\%  & 43.61\% & 78.31\% & 91.09\% & 84.70\% \\
VCD      & 76.64\% & 10.36\% & 43.50\% & 76.64\% & 89.64\% & 83.14\% \\
ICD      & 72.91\% & 7.02\%  & 39.97\% & 72.91\% & 92.98\% & 82.94\% \\
AvisC    & 76.02\% & 7.04\%  & 41.53\% & 76.02\% & 92.96\% & 84.49\% \\
\bottomrule
\end{tabular}
\end{table}

\begin{figure*}[t]
    \centering
    \includegraphics[width=\textwidth]{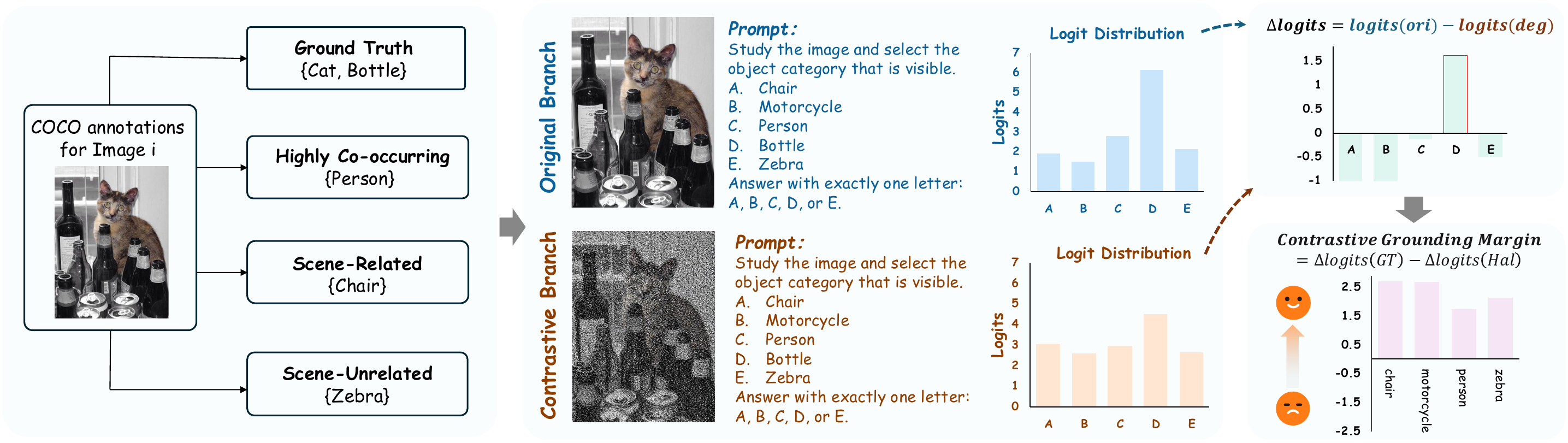}
    \caption{Overview of \concor{} and CGM. Ground-truth and hallucinated candidates are constructed from COCO annotations and co-occurrence statistics and evaluated under the same prompt. Logits from the original and contrastive branches are used to compute the contrastive signal $\Delta_w$, from which CGM measures the change in the relative logit advantage of ground-truth over hallucinated candidates.}
    \label{fig:contrastive_signal_benchmark}
\end{figure*}

\section{CONCOR}
\label{sec:method}

The preceding observations show that existing evaluation scores can be affected by decoding constraints, generation context, and response bias, making it difficult to isolate the corrective signal provided by the contrastive branch. 
Evaluating this signal therefore requires comparable object-decision conditions under which we can directly measure whether contrastive intervention increases the relative advantage of ground-truth over hallucinated candidates.
We therefore introduce \concor{} (\underline{Con}trastive \underline{Cor}rection Evaluation), a benchmark dataset that places ground-truth and hallucinated objects in the same multiple-choice candidate set, fixes the generation context and decoding position, and extracts raw logits from both branches without candidate filtering. 
Within \concor{}, we define the \textbf{Contrastive Grounding Margin (CGM)} to measure the direction and magnitude of the resulting change in relative advantage (summarized in Figure~\ref{fig:contrastive_signal_benchmark}).

\subsection{Benchmark Design}
\label{sec:benchmark_design}
We sample 2,000 distinct images from COCO Val2014~\cite{lin2014microsoft}, with balanced coverage across its 80 object categories, and construct a five-option object-recognition question for each image. COCO instance annotations determine the ground-truth candidates, while unannotated categories form the hallucinated candidate pool. Single-answer questions contain one ground-truth and four hallucinated candidates. Multiple-answer questions contain two to four ground-truth candidates, with the remaining filled by hallucinated candidates.

To evaluate ground-truth objects against different types of hallucinated candidates, we construct candidate sets that pair ground-truth objects with hallucinated objects selected using different co-occurrence criteria. 
Let $P_I$ denote the set of categories annotated in image $I$.
For an annotated category $a$ and an unannotated category $o$, we define their conditional co-occurrence as 
\begin{equation}
    C(a,o)=\frac{N(a,o)}{N(a)},
\end{equation}
where $N(a)$ is the number of COCO images containing $a$, and $N(a,o)$ is the number containing both $a$ and $o$.
For each unannotated candidate $o\notin P_I$, we then define
\begin{equation}
S_{\mathrm{scene}}(o,I)
= \frac{1}{|P_I|}
   \sum_{a \in P_I} C(a,o),
   \end{equation}
\begin{equation}
S_{\max}(o,I)
= \max_{a \in P_I} C(a,o).
\end{equation}
These scores measure the candidate's average and strongest co-occurrence with the annotated categories, respectively.
We use them to construct four types of hallucinated candidates---a highly co-occurring candidate that maximizes $S_{\max}$,
a scene-related candidate that maximizes $S_{\mathrm{scene}}$,
a scene-unrelated candidate that minimizes $S_{\mathrm{scene}}$,
and a random candidate sampled uniformly from the remaining unannotated categories.
This construction exposes each ground-truth object to hallucinated alternatives with different co-occurrence profiles.
An illustrative example of the candidate construction process is provided in Appendix~\ref{appendix: didate Construction Example}.

\subsection{Contrastive Grounding Margin}
\label{sec:CGM_definition}
\subsubsection{Definition}
Under the controlled setting of \concor{}, let $\ell_o(w)$ and $\ell_d(w)$ denote the logits assigned to candidate $w$ by the original and contrastive branches, respectively. For each candidate, we define its contrastive signal as
\begin{equation}
\Delta_w = \ell_o(w) - \ell_d(w).
\label{eq:cgm_signal}
\end{equation}

For a ground-truth candidate $g$ and a hallucinated candidate $h$,
we define the \textbf{Contrastive Grounding Margin (CGM)} as
\begin{equation}
\mathrm{CGM}(g,h) = \Delta_g - \Delta_h.
\label{eq:cgm_pairwise}
\end{equation}
CGM measures whether the contrastive signal favors the ground-truth candidate over the hallucinated candidate, and by how much.
This interpretation follows directly from the  contrastive decoding rule, as shown in Eq.~\ref{eq:cd}, 
where $\alpha > 0$ controls the contrast strength.
Based on Eq.~\ref{eq:pairwise_margin} and Eq.~\ref{eq:cgm_pairwise}, the resulting change in the logit margin is therefore $\alpha\,\mathrm{CGM}(g,h)$.
A positive CGM means that the contrastive intervention increases the relative advantage of the ground-truth candidate, a negative CGM means that it increases the relative advantage of the hallucinated candidate,
and a zero CGM leaves their relative margin unchanged. The magnitude of CGM indicates the strength of this effect, while the actual change in the logit margin is scaled by $\alpha$.

To accommodate both single-answer and multiple-answer settings, let $G$ and $H$ denote the sets of ground-truth and hallucinated candidates in a question. We define question-level CGM as
\begin{equation}
\mathrm{CGM}(G,H)
=
\frac{1}{|G|}\sum_{g \in G}\Delta_g
-
\frac{1}{|H|}\sum_{h \in H}\Delta_h.
\label{eq:cgm_question}
\end{equation}

This equals the mean pairwise CGM over all ground-truth--hallucinated candidate pairs and captures the average change in the relative logit advantage of ground-truth over hallucinated candidates per unit contrast strength.

\subsubsection{Diagnostic Properties}
\label{sec:diagnostic_prop}
Beyond quantifying the overall strength of contrastive correction, CGM provides a direct characterization of how a contrastive intervention changes the competition between ground-truth and hallucinated candidates. Its formulation yields three diagnostic properties concerning \emph{directional consistency}, \emph{correction composition}, and \emph{correction reliability}.

\vspace{-3mm}
\paragraph{Property 1: Directional consistency.}
For each question, the sign of $\mathrm{CGM}(G,H)$ indicates whether the contrastive intervention increases or decreases the average relative advantage. Across a benchmark containing $N$ questions, we define the positive-CGM rate as
\begin{equation}
R_{+}
=
\frac{1}{N}
\sum_{i=1}^{N}
\mathbb{I}
\left[
\mathrm{CGM}(G_i,H_i)>0
\right].
\end{equation}
The positive-CGM rate therefore measures how consistently a CD method produces adjustments in the desired direction across samples. A high rate indicates that the contrastive intervention usually strengthens the relative advantage of ground-truth candidates, whereas a low rate indicates that the intervention is frequently misdirected.

\vspace{-3mm}
\paragraph{Property 2: Correction composition.}
While CGM measures the net change in the relative advantage of ground-truth over hallucinated candidates, this change can arise from different candidate-level adjustments. To characterize its composition, we separately examine the contrastive signals of the ground-truth candidates, the highest-scoring hallucinated candidate under the original branch, and the remaining hallucinated candidates, denoted by $\Delta_{\mathrm{GT}}$, $\Delta_{\mathrm{MaxHal}}$, and $\Delta_{\mathrm{OtherHal}}$, respectively. Their relative values reveal whether the correction arises primarily from strengthening ground-truth candidates, suppressing the strongest hallucinated competitor, suppressing hallucinated candidates more broadly, or a combination of these effects. Thus, CGM together with its candidate-level decomposition distinguishes contrastive interventions that produce similar net corrections through different underlying adjustment patterns.

\vspace{-3mm}
\paragraph{Property 3: Correction reliability.}
Let $M_o(g,h)=\ell_o(g)-\ell_o(h)$
denote the original margin between a ground-truth candidate $g$ and a hallucinated candidate $h$. Under contrastive decoding, the corresponding margin becomes
\begin{equation}
M_{\mathrm{CD}}(g,h)
=M_o(g,h)+\alpha \mathrm{CGM}(g,h).
\end{equation}
Thus, CGM measures the change in the ground-truth-versus-hallucination margin per unit contrast strength.
A positive CGM can reverse an initially negative margin when the resulting adjustment is sufficiently large, correcting an incorrect candidate ordering. Conversely, a negative CGM can overcome an initially positive margin and turn a correct ordering into an incorrect one. We therefore examine incorrect-to-correct (W$\rightarrow$C) and correct-to-incorrect (C$\rightarrow$W) prediction transitions to characterize the beneficial and harmful consequences of contrastive adjustments.

\section{Experiments}
\label{sec:experiments}

We conduct experiments to evaluate existing CD methods through \concor{} and investigate how the diagnostic properties of CGM manifest in practice. Specifically, we organize our experiments around three research questions:
\begin{itemize}
    \item \textbf{RQ1.} \textit{How effective are existing CD methods, and what contrastive correction patterns underlie their performance?}
    \item \textbf{RQ2.} \textit{How does contrast strength affect these corrections and prediction performance?}
    \item \textbf{RQ3.} \textit{Are the observed correction patterns consistent across models and task settings?}
\end{itemize}

\subsection{Setup}
We compare baseline decoding with VCD~\cite{leng2024mitigating}, ICD~\cite{wang2024mitigating}, SID~\cite{huo2025self}, and AvisC~\cite{woo2025don} on \concor, covering the 7B and 13B variants of LLaVA-1.5~\cite{Liu_2024_CVPR} and InstructBLIP-Vicuna~\cite{dai2023instructblip}, as well as single-answer and multiple-answer settings. A brief introduction of the CD methods are provided in Appenix~\ref{appendix:cd_method_intro}.
For each CD method, we extract raw logits from the original and contrastive branches at the first answer position before candidate filtering.
We report mean CGM, the positive-CGM rate to measure the magnitude and directional consistency of the contrastive correction, respectively, together with image-level sampling accuracy to measure its effect on final predictions.

\begin{table*}[t]
\centering
\caption{CGM and image-level accuracy under single- and multiple-ground-truth settings across different models and model sizes.}
\label{tab:hss_gt_settings}

\footnotesize
\setlength{\tabcolsep}{3.0pt}
\renewcommand{\arraystretch}{1.15}

\begin{tabular*}{\textwidth}{
    @{\extracolsep{\fill}}
    ll*{12}{c}
    @{}
}
\toprule
& &
\multicolumn{3}{c}{LLaVA-1.5-7B} &
\multicolumn{3}{c}{LLaVA-1.5-13B} &
\multicolumn{3}{c}{InstructBLIP-Vicuna-7B} &
\multicolumn{3}{c}{InstructBLIP-Vicuna-13B} \\
\cmidrule(lr){3-5}
\cmidrule(lr){6-8}
\cmidrule(lr){9-11}
\cmidrule(lr){12-14}

Setting & Method
& CGM & Positive & Acc.
& CGM & Positive & Acc.
& CGM & Positive & Acc.
& CGM & Positive & Acc. \\
\midrule

\multirow{5}{*}{\rotatebox[origin=c]{90}{Single GT}}
& Baseline
& ---    & ---    & 71.1\%
& ---    & ---    & 87.2\%
& ---    & ---    & 47.2\%
& ---    & ---    & 21.8\% \\

& VCD
& -0.029 & 41.6\% & 69.2\%
&  0.178 & 60.8\% & 85.1\%
& -0.078 & 33.9\% & 41.2\%
& -0.025 & 40.9\% & 20.6\% \\

& ICD
&  0.393 & 93.6\% & 77.5\%
&  0.120 & 83.3\% & 87.7\%
& -0.059 & 42.4\% & 41.7\%
& -0.370 &  5.6\% & 12.8\% \\

& SID
&  1.279 & 74.7\% & 81.5\%
&  4.151 & 97.0\% & 93.0\%
&  0.357 & 75.1\% & 50.1\%
&  0.491 & 88.3\% & 31.6\% \\

& AvisC
&  3.254 & 98.3\% & 83.1\%
&  4.683 & 98.5\% & 91.0\%
&  1.525 & 97.2\% & 71.3\%
&  0.760 & 92.0\% & 39.8\% \\

\midrule

\multirow{5}{*}{\rotatebox[origin=c]{90}{Multiple GT}}
& Baseline
& ---    & ---    & 88.0\%
& ---    & ---    & 82.0\%
& ---    & ---    & 87.0\%
& ---    & ---    & 52.5\% \\

& VCD
& -0.030 & 46.5\% & 87.5\%
&  0.165 & 64.5\% & 79.0\%
&  0.004 & 50.5\% & 85.5\%
&  0.061 & 64.5\% & 52.5\% \\

& ICD
&  0.115 & 79.5\% & 91.0\%
&  0.075 & 84.0\% & 81.0\%
&  0.043 & 67.0\% & 82.0\%
&  0.074 & 86.0\% & 55.0\% \\

& SID
&  0.545 & 73.5\% & 91.0\%
&  1.836 & 97.0\% & 93.5\%
&  0.170 & 72.5\% & 92.5\%
&  0.836 & 97.5\% & 64.5\% \\

& AvisC
&  2.131 & 99.5\% & 94.0\%
&  2.267 & 99.0\% & 93.5\%
&  0.859 & 94.5\% & 96.0\%
&  0.996 & 98.0\% & 89.5\% \\

\bottomrule
\end{tabular*}

\end{table*}

\subsection{RQ1: How Effective Are Existing CD Methods?}
\label{sec:effectiveness}
We first evaluate existing CD methods under the main setting of LLaVA-1.5-7B with single-answer questions. As shown in Table~\ref{tab:hss_gt_settings}, Baseline decoding achieves an accuracy of 71.1\%. The CD methods exhibit substantial differences in both prediction accuracy and their underlying contrastive corrections. VCD obtains a mean CGM of $-0.029$ and a positive-CGM rate of 41.6\%, with accuracy decreasing to 69.2\%. In contrast, ICD, SID, and AvisC achieve positive mean CGMs of $0.393$, $1.279$, and $3.254$, respectively, and improve accuracy to 77.5\%, 81.5\%, and 83.1\%. These results show that the performance differences among CD methods are associated with substantial differences in the contrastive corrections they produce.

\vspace{-3mm}
\paragraph{Directional consistency.}
\textbf{CD methods differ substantially in the directional consistency of their contrastive adjustments.}
Table~\ref{tab:transition_logit_changes1} shows that VCD has a positive-CGM rate of only $41.6\%$, suggesting that its contrastive adjustments decrease rather than increase the ground-truth advantage on most samples.
In contrast, ICD and AvisC achieve positive-CGM rates of $93.6\%$ and $98.3\%$, respectively, indicating highly consistent correction toward the ground truth.
SID falls between these methods, with a positive-CGM rate of $74.7\%$: although its adjustments favor ground-truth candidates in most cases, they are less consistent than those of ICD and AvisC.
These differences help distinguish methods that reliably apply corrections in the desired direction from those whose corrections are frequently misdirected.

\vspace{-3mm}
\paragraph{Correction composition.}
\textbf{CD methods achieve positive correction through different candidate-level adjustment patterns.}
On successfully corrected samples, ICD makes small but relatively uniform adjustments to hallucinated candidates, decreasing the logits of the strongest and remaining hallucinations by $0.10$ and $0.10$, respectively.
SID instead concentrates its adjustments on separating the ground truth from the strongest hallucination, raising the former by $2.05$ and lowering the latter by $0.64$, while the remaining hallucinations change by only $+0.03$.
AvisC exhibits a different pattern: it increases the logits of all three candidate groups by $17.44$, $16.16$, and $14.49$. Because these increases are progressively smaller for the strongest and remaining hallucinations, AvisC increases the ground-truth margin over these groups by $1.28$ and $2.95$.
These results show that positive contrastive correction can arise through different mechanisms, including direct suppression of hallucinated candidates or a larger relative increase in ground-truth logits, even when hallucinated logits themselves increase.

\vspace{-3mm}
\paragraph{Correction reliability.}
\textbf{CD methods differ substantially in whether their contrastive adjustments correct existing errors without introducing new ones.}
As shown in Table~\ref{tab:transition_logit_changes1}, SID corrects the most samples ($288$) but also introduces $81$ errors, compared with $8$ for ICD and $15$ for AvisC.
Table~\ref{tab:transition_logit_changes2} further shows that, on SID's W$\rightarrow$C samples, the ground-truth logit increases while the strongest hallucination decreases, increasing their margin by $2.69$.
On C$\rightarrow$W samples, this pattern reverses: the ground-truth logit decreases by $0.76$, while the hallucinated candidate that becomes the final prediction increases by $1.52$, producing a margin change of $-2.28$.
This adverse margin change exceeds those of ICD ($-0.35$) and AvisC ($-0.81$).
When the reverse adjustment exceeds the ground truth's initial logit lead, the hallucination overtakes it, turning a correct prediction into an error.

\begin{table}[t]
\centering
\caption{
Candidate-level contrastive signals and prediction transitions.
}
\label{tab:transition_logit_changes1}
\footnotesize
\setlength{\tabcolsep}{0pt}
\renewcommand{\arraystretch}{1.1}

\begin{tabular*}{\columnwidth}{
    @{\extracolsep{\fill}}
    lcccccc
    @{}
}
\toprule
Method
& $\Delta_{\mathrm{GT}}$
& $\Delta_{\mathrm{MaxHal}}$
& $\Delta_{\mathrm{OtherHal}}$
& $\mathrm{C}\!\rightarrow\!\mathrm{W}$
& $\mathrm{W}\!\rightarrow\!\mathrm{C}$
& Positive-CGM \\
\midrule
Baseline
& \textemdash
& \textemdash
& \textemdash
& \textemdash
& \textemdash
& \textemdash \\

VCD
& $+1.13$
& $-0.27$
& $-0.17$
& 100
& 62
& 41.6\% \\

ICD
& $+0.54$
& $-0.10$
& $-0.10$
& \textbf{8}
& 135
& 93.6\% \\

SID
& $\mathbf{+2.05}$
& $\mathbf{-0.64}$
& $+0.03$
& 81
& \textbf{288}
& 74.7\% \\

AvisC
& $+17.44$
& $+16.16$
& $+14.49$
& 15
& 255
& 98.3\% \\
\bottomrule
\end{tabular*}
\end{table}

\begin{table}[t]
\centering
\caption{
Candidate-level contrastive signals for prediction transitions. 
}
\label{tab:transition_logit_changes2}
\footnotesize
\setlength{\tabcolsep}{0pt}
\renewcommand{\arraystretch}{1.1}

\begin{tabular*}{\columnwidth}{
    @{\extracolsep{\fill}}
    lccccc
    @{}
}
\toprule
Method
& Transition
& $\Delta_{\mathrm{GT}}$
& $\Delta_{\mathrm{MaxHal}}$
& $\Delta_{\mathrm{Margin}}$
& Acc. \\
\midrule

\multirow{2}{*}{SID}
& $\mathrm{W}\!\rightarrow\!\mathrm{C}$
& $+2.05$
& $-0.64$
& $\mathbf{+2.69}$
& \multirow{2}{*}{81.5\%} \\
& $\mathrm{C}\!\rightarrow\!\mathrm{W}$
& $-0.76$
& $+1.52$
& $\mathbf{-2.28}$
& \\
\addlinespace[2pt]

\multirow{2}{*}{ICD}
& $\mathrm{W}\!\rightarrow\!\mathrm{C}$
& $+0.54$
& $-0.10$
& $\mathbf{+0.64}$
& \multirow{2}{*}{77.5\%} \\
& $\mathrm{C}\!\rightarrow\!\mathrm{W}$
& $-0.04$
& $+0.31$
& $\mathbf{-0.35}$
& \\
\addlinespace[2pt]

\multirow{2}{*}{AvisC}
& $\mathrm{W}\!\rightarrow\!\mathrm{C}$
& $+17.44$
& $+16.16$
& $\mathbf{+1.28}$
& \multirow{2}{*}{83.1\%} \\
& $\mathrm{C}\!\rightarrow\!\mathrm{W}$
& $+15.98$
& $+16.79$
& $\mathbf{-0.81}$
& \\

\bottomrule
\end{tabular*}
\vspace{-18pt}
\end{table}

\begin{table}[t]
\centering
\caption{Effect of Contrastive Decoding Strength on Accuracy for LLaVA-1.5-7B.}
\label{tab:contrastive_strength_accuracy}
\footnotesize
\renewcommand{\arraystretch}{1.1}

\begin{tabular*}{\columnwidth}{
    @{\extracolsep{\fill}}
    lcccc
    @{}
}
\toprule
Method
& CGM
& Positive
& Acc. ($\alpha=1$)
& Acc. ($\alpha=3$) \\
\midrule
Baseline & \textemdash & \textemdash & 71.1\% & 71.1\% \\
VCD      & $-0.029$ & 41.6\% & 69.2\% & 63.3\% \\
ICD      & $0.393$  & 93.6\% & 77.5\% & 83.2\% \\
SID      & $1.279$  & 74.7\% & 81.5\% & 77.2\% \\
AvisC    & $3.254$  & 98.3\% & 83.1\% & 87.2\% \\
\bottomrule
\end{tabular*}
\vspace{-15pt}
\end{table}

\begin{figure*}[t]
    \centering
    \includegraphics[width=0.95\textwidth]{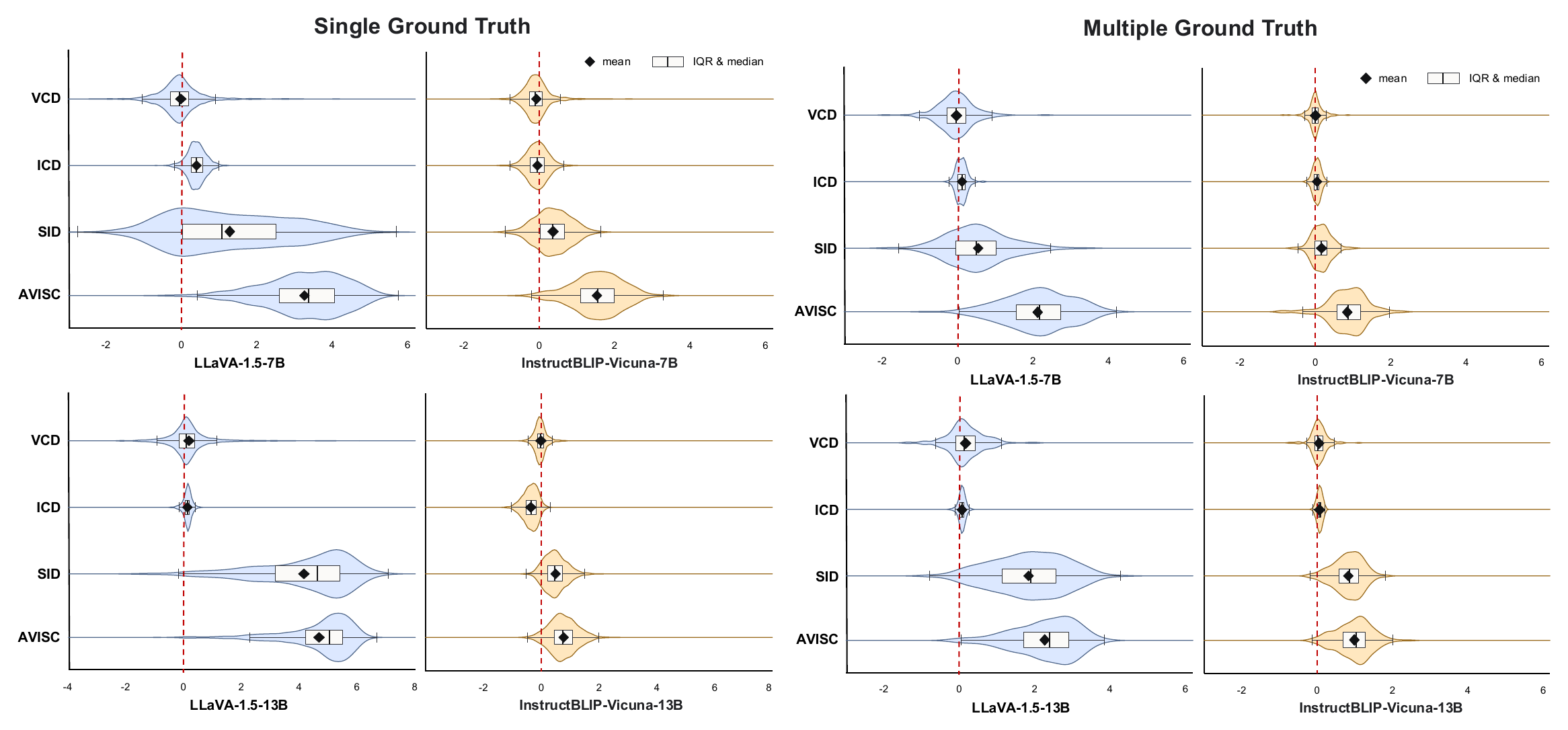}
    \caption{Distribution of CGM across models and ground-truth settings. Violin plots show the CGM distributions of VCD, ICD, SID, and AvisC for LLaVA-1.5 and InstructBLIP-Vicuna at 7B and 13B scales under single- and multiple-ground-truth settings. Diamonds indicate means, boxes show the interquartile range with the median, and whiskers extend to 1.5 times the interquartile range.}
    \label{fig:branch_signal_violin}
\end{figure*}

\subsection{RQ2: How Does Contrast Strength Affect CD Methods?}
\label{sec:rq_contrastive_strength}
The correction patterns identified in RQ1 suggest that CD methods may respond differently to increased contrast strength. Under the CD formulation, increasing $\alpha$ scales the contrastive adjustment without changing its direction. Stronger contrast should therefore amplify beneficial adjustments when CGM is positive, but can also amplify harmful adjustments when CGM is negative. We examine this relationship by increasing $\alpha$ from $1$ to $3$.

Table~\ref{tab:contrastive_strength_accuracy} shows that increasing $\alpha$ from $1$ to $3$ raises ICD's accuracy from $77.5\%$ to $83.2\%$ and AvisC's from $83.1\%$ to $87.2\%$.
As shown in RQ1, these methods have positive-CGM rates of $93.6\%$ and $98.3\%$, respectively, indicating that their adjustments favor ground-truth candidates on most samples.
Increasing the contrast strength therefore predominantly amplifies beneficial adjustments.
For ICD in particular, amplifying its initially mild but directionally reliable adjustments yields a gain of $5.7$ percentage points.
In contrast, SID has a lower positive-CGM rate of $74.7\%$.
Although SID produces strong beneficial corrections, RQ1 also shows that its adverse adjustments can be sufficiently large to reverse initially correct predictions. 
Increasing $\alpha$ can amplify these adjustments as well, and its  accuracy decreases from $81.5\%$ to $77.2\%$.
VCD further illustrates the risk of amplifying an unreliable contrastive direction: its mean CGM is $-0.029$, its positive-CGM rate is only $41.6\%$, and its accuracy falls from $69.2\%$ to $63.3\%$.
These results demonstrate that CGM offers more than a comparison of final performance: it reveals how different CD methods suppress hallucinations and under what conditions they may amplify errors.
This analysis can inform method selection and decoding configuration while identifying opportunities to improve contrastive branches and strategies, supporting more effective CD design and application.

\subsection{RQ3: Are the Observed Effects Consistent Across Models and Task Settings?}
\label{sec:rq_robustness}

\textbf{The main differences among CD methods remain consistent across model architectures, parameter scales, and question formats.}
Table~\ref{tab:hss_gt_settings} shows that, when LLaVA-1.5-7B is replaced with InstructBLIP-Vicuna-7B, AvisC and SID achieve mean CGM values of $1.525$ and $0.357$ in the single-answer setting, still exceeding VCD's $-0.078$ and ICD's $-0.059$.
The same ordering is observed at a larger model scale: on LLaVA-1.5-13B, AvisC and SID achieve mean CGM values of $4.683$ and $4.151$, compared with $0.178$ for VCD and $0.120$ for ICD.
These results are aligned with the finding that feature-level methods provide stronger corrective signals across models.

As shown in Figure~\ref{fig:branch_signal_violin}, this consistency extends to multiple-answer tasks, where multiple ground-truth candidates jointly compete with hallucinated candidates.
On LLaVA-1.5-7B, AvisC and SID achieve mean CGM values of $2.131$ and $0.545$, respectively, again exceeding ICD's $0.115$ and VCD's $-0.030$.
Across all eight settings comprising four model variants and two question formats in Table~\ref{tab:hss_gt_settings}, AvisC consistently achieves a higher mean CGM than SID, and both exceed the ICD and VCD.
Additional analyses further confirm the robustness of these findings across different option-selection settings (Appendix~\ref{appendix: across options settings}), levels of annotation missingness (Appendix~\ref{sec:coco-missingness-robustness}), and dataset construction (Appendix~\ref{appendix: across datasets}).
These results show that the main differences in contrastive correction identified by CGM are robust across the tested model architectures, parameter scales, and question formats.

\section{Conclusion}
We revisit the evaluation of contrastive decoding for mitigating object hallucinations in multimodal large language models.
We show that existing generative and discriminative evaluations can be confounded by decoding constraints, changes in generation trajectories, and response biases.
To address these limitations while complementing existing evaluations, we introduce \concor, a benchmark that directly evaluates the competition between ground-truth and different types of hallucinated candidates, together with Contrastive Grounding Margin (CGM), which measures how contrastive decoding changes their relative logit margins. 
Beyond measuring the overall strength of contrastive correction, CGM provides diagnostics of its directional consistency, correction composition, and reliability. Applying this framework to existing CD methods reveals substantial differences in how they produce corrections.
Looking forward, these diagnostics provide several directions for improving CD methods. For example, future work could develop contrastive branches that produce more consistently beneficial corrections, adapt contrast strength according to the direction and magnitude of the contrastive signal, or explicitly control how adjustments are distributed between ground-truth and hallucinated candidates. 
Together, \concor{} and CGM provide a more controlled and interpretable framework for evaluating contrastive decoding and understanding how its underlying adjustments contribute to hallucination mitigation.

\section*{LLM Usage Statement}

ChatGPT was used to support the organization and development of the manuscript for clarity. It also assisted with preliminary reference discovery by identifying potentially relevant literature for further review.

{\small 
\bibliographystyle{ieeenat_fullname}
\bibliography{main}
}
\appendix

\section{Details of \concor}
\label{appendix:benchmark_details}

\begin{figure*}[t]
    \centering
    \includegraphics[width=\textwidth]{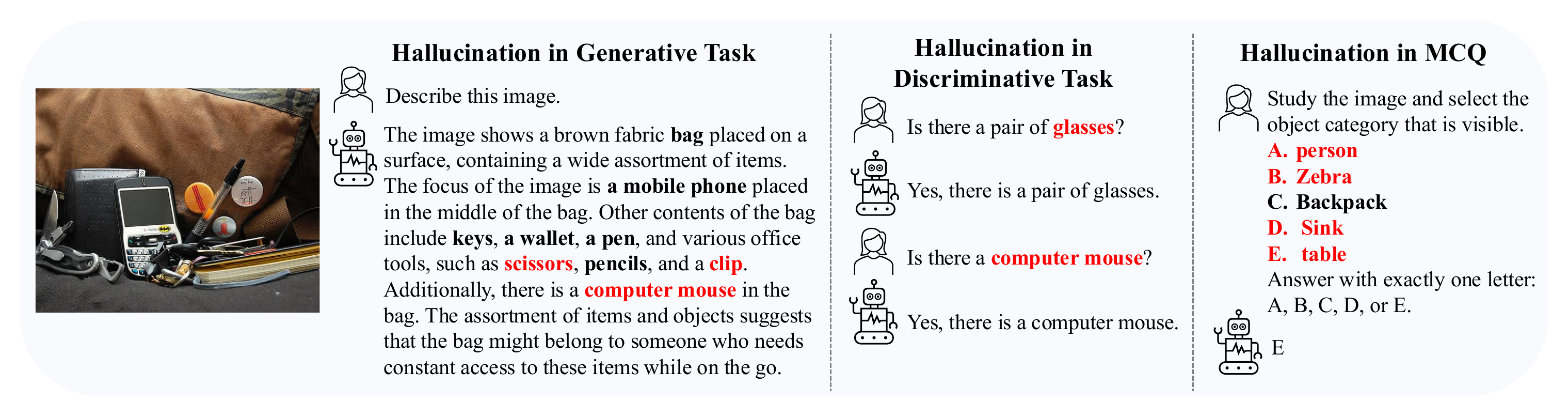}
    \caption{Comparison of hallucination evaluation in generative, discriminative, and multiple-choice question (MCQ) tasks.}
    \label{fig:benchmark_comparison}
\end{figure*}

Figure~\ref{fig:benchmark_comparison} compares hallucination evaluation under generative, discriminative, and MCQ formulations. \concor{} uses a controlled MCQ setting in which ground-truth and hallucinated candidates directly compete within the same question and answer context.

\subsection{Dataset Construction}

\concor{} can be instantiated on datasets with image-level object-category annotations. Let $\mathcal{C}$ denote the category vocabulary, $P_I$ the categories annotated as present in image $I$, and $U_I=\mathcal{C}\setminus P_I$ the  categories not annotated as present. Ground-truth candidates are selected from $P_I$, while hallucinated candidates are constructed from $U_I$.

Our main experiments instantiate \concor{} on the COCO 2014~\cite{lin2014microsoft} validation set using its images, 80 object categories, and instance annotations.

\subsubsection{Single-Ground-Truth Setting}

The single-ground-truth setting contains 2,000 distinct image questions, sampled to provide category-aware coverage over the 80 COCO object categories. Each question contains one ground-truth candidate and four hallucinated candidates representing different co-occurrence characteristics. The model receives the following prompt:

\begin{quote}
\footnotesize\ttfamily
Study the image and select the object category that is visible.\\[2pt]
A. \{object\_A\}\\
B. \{object\_B\}\\
C. \{object\_C\}\\
D. \{object\_D\}\\
E. \{object\_E\}\\[2pt]
Answer with exactly one letter: A, B, C, D, or E.
\end{quote}

The four hallucinated candidates are constructed using the co-occurrence criteria described in Sec.~\ref{sec:benchmark_design}. They are selected sequentially without replacement. Ties are resolved using the secondary co-occurrence score and then the global category frequency.

To reduce candidate-position bias, we generate five cyclic A--E permutations for each question, so that every semantic candidate appears exactly once at each answer position. This produces $2{,}000\times5=10{,}000$ model inputs.

\subsubsection{Multiple-Ground-Truth Setting}

The multiple-ground-truth setting contains 200 image questions. Each question has five candidates, of which two to four are ground-truth candidates and the remainder are hallucinated candidates. The model receives the following prompt:

\begin{quote}
\footnotesize\ttfamily
What objects are in this image? Select every option whose object category is visible.\\[2pt]
A. \{object\_A\}\\
B. \{object\_B\}\\
C. \{object\_C\}\\
D. \{object\_D\}\\
E. \{object\_E\}\\[2pt]
Answer with all correct letters in alphabetical order, separated by commas (for example: A, C). Select all and only the visible options. Do not output anything else.
\end{quote}

Because each question contains five candidates, the number of hallucinated candidates satisfies $|H|=5-|G|$. The dataset contains 67 questions with $(|G|,|H|)=(2,3)$, 67 with $(3,2)$, and 66 with $(4,1)$.

Ground-truth candidates are selected from annotated objects occupying at least $0.2\%$ of the image area and are ordered by annotated area. Larger objects are prioritized. 
Hallucinated candidates are selected from unannotated categories in descending order of $S_{\mathrm{scene}}$. Ties are resolved using $S_{\max}$ and then the global category frequency.

As in the single-ground-truth setting, each question is converted into five cyclic permutations, producing $200\times5=1{,}000$ model inputs.

\subsection{Metric Computation}

We use the candidate-level contrastive adjustment and Contrastive Grounding Margin (CGM) defined in Sec.~\ref{sec:CGM_definition}. For each semantic candidate, its contrastive adjustment is first averaged across the five cyclic permutations to alleviate candidate-position effects. These position-averaged adjustments are then used to compute the question-level CGM following Eq.~\ref{eq:cgm_question}.

\paragraph{Single-ground-truth accuracy.}

A permutation is correct when the predicted label matches the unique ground-truth label. An image question is counted as correct if at least three of its five permutations are correct.

\paragraph{Multiple-ground-truth accuracy.}

We use a no-false-positive criterion. A permutation is correct when its predicted set is nonempty and is a subset of the ground-truth set. Thus, missing some ground-truth candidates is permitted, whereas selecting any hallucinated candidate makes the prediction incorrect. An image question is counted as correct if at least three of its five permutations satisfy this criterion.

\section{Contrastive Decoding for Hallucinations}
\label{appendix:cd_method_intro}
This section introduces four methods for mitigating hallucinations in multimodal large language models: Visual Contrastive Decoding (VCD), Instruction Contrastive Decoding (ICD), Self-Introspective Decoding (SID), and Attentional Vision Calibration (AvisC). These methods construct a hallucination-amplified reference branch at inference time and contrast its output distribution with that of the original branch to adjust next-token scores.

\subsection{General Formulation}

Given an image $I$, an instruction $x$, and previously generated tokens $y_{<t}$, the original logits at decoding step $t$ are

\[
\mathbf{z}_t
=
f_{\theta}(I,x,y_{<t}).
\]

Contrastive decoding additionally constructs a perturbed condition $c^{-}$ and computes the corresponding reference logits:

\[
\mathbf{z}_t^{-}
=
f_{\theta}(c^{-},y_{<t}).
\]

The contrastive logits can generally be written as

\[
\mathbf{z}_t^{\mathrm{CD}}
=
(1+\alpha)\mathbf{z}_t
-
\alpha\mathbf{z}_t^{-},
\]

where $\alpha$ controls the contribution of the reference branch.
Some methods additionally apply a plausibility constraint that restricts decoding to high-probability candidates under the original
distribution.

\subsection{Visual Contrastive Decoding}

VCD constructs its reference branch by perturbing the visual input.
Gaussian noise is progressively added to the original image $I$, producing a distorted image $\widetilde{I}$. The model then computes

\[
\mathbf{z}_t
=
f_{\theta}(I,x,y_{<t}),
\qquad
\widetilde{\mathbf{z}}_t
=
f_{\theta}(\widetilde{I},x,y_{<t}).
\]

VCD uses the logits conditioned on the original image as the main branch and those conditioned on the distorted image as the reference branch. The two sets of logits are combined through weighted subtraction. A plausibility constraint then removes candidates with low probabilities under the original distribution, and the next token is selected from the remaining candidate set.

\subsection{Instruction Contrastive Decoding}

ICD constructs its reference branch by perturbing the textual instruction. A predefined disturbance prefix or role description is added to the original instruction to obtain a disturbed instruction $\widetilde{x}$. The model computes

\[
\mathbf{z}_t
=
f_{\theta}(I,x,y_{<t}),
\qquad
\widetilde{\mathbf{z}}_t
=
f_{\theta}(I,\widetilde{x},y_{<t}).
\]

At each decoding step, ICD subtracts the logits conditioned on the disturbed instruction from those conditioned on the original instruction. A contrastive coefficient controls the magnitude of the subtraction. A high-probability candidate set is then constructed from the original distribution, and token normalization and selection are performed within this set.

\subsection{Self-Introspective Decoding}

SID constructs its reference branch using the model's internal attention information. Its main component is the Context and Text-aware Token Selection ($CT^2S$) strategy, which identifies visual tokens receiving relatively low attention from the current generation position. At an early decoder layer $i$, $CT^2S$ computes the importance score for each visual token $v_j$ by averaging the attention from the current generation position across all attention heads:

\[
s_i(v_j)
=
\frac{1}{H}
\sum_{h=1}^{H}
\mathbf{A}_i^{(h)}[t,j],
\]

where $H$ denotes the number of attention heads and $\mathbf{A}_i^{(h)}[t,j]$ denotes the attention assigned by the current token position $t$ to visual token $v_j$ in head $h$. The visual tokens with the lowest importance scores are selected as

\[
\mathcal{V}_i^{\downarrow}
=
\operatorname{BottomK}
\left(
\left\{v_j\right\}_{j=1}^{N};
\left\{s_i(v_j)\right\}_{j=1}^{N}
\right).
\]

The reference branch retains visual tokens with low importance scores and passes them to the subsequent decoder layers to obtain the reference logits. In parallel, the original branch uses the complete set of visual tokens to produce the original logits. 
SID then performs weighted subtraction between the two branches and generates the next token from the calibrated distribution. Visual-token selection is dynamically repeated at different steps of autoregressive generation.

\subsection{Attentional Vision Calibration}

AvisC first identifies blind tokens from the model's internal visual-attention distributions. It aggregates visual attention across selected decoder layers and attention heads to obtain an attention score $A_j$ for each visual token $j$. Let $\mu_A$ and $\sigma_A$ denote the mean and standard deviation of these scores. Tokens satisfying the following condition are identified as blind tokens:

\[
\mathcal{B}
=
\left\{
j
\;\middle|\;
A_j>\mu_A+\lambda\sigma_A
\right\},
\]
where $\lambda$ controls the threshold for blind-token selection.

AvisC then constructs a biased visual representation $\mathcal{V}^{*}$ by retaining the visual tokens indexed by $\mathcal{B}$ while setting the remaining visual tokens to zero. 
Given the original visual representation $\mathcal{V}$, the model computes
\[
\boldsymbol{\ell}_t
=
\log p
\left(
y_t
\mid
\mathcal{V},x,y_{<t}
\right),
\qquad
\boldsymbol{\ell}_t^{*}
=
\log p
\left(
y_t
\mid
\mathcal{V}^{*},x,y_{<t}
\right),
\]
where $\boldsymbol{\ell}_t$ and $\boldsymbol{\ell}_t^{*}$ denote the next-token logits obtained from the original and biased visual representations, respectively. AvisC contrasts these logits to adjust the next-token scores, using the biased representation as the reference branch.

\section{Additional Experimental Details}

\subsection{Computational Resources}
All model-inference experiments are run on a single NVIDIA A800-SXM4 80GB, NVIDIA RTX 6000D, NVIDIA GeForce RTX 5090, or NVIDIA GeForce RTX 4090 GPU, depending on the experimental group. Within each experimental group, all compared methods and settings are evaluated using the same GPU model to ensure consistent hardware conditions.

\subsection{Main Experiments}
\label{appendix:main_exp_details}

\subsubsection{Models and General Setup}

We evaluate VCD, ICD, SID, and AvisC on LLaVA-1.5-7B, LLaVA-1.5-13B, InstructBLIP-Vicuna-7B, and InstructBLIP-Vicuna-13B, together with standard decoding as the baseline (denoted as Baseline).

All models are used for inference without additional training or fine-tuning. For each model and sample, all methods use the same image, question, candidate set, and candidate ordering. The CD methods use the same original-branch computation and differ only in how the reference branch is constructed. Baseline uses only the original branch.

LLaVA uses the \texttt{llava\_v1} conversation template, while InstructBLIP uses the official \texttt{InstructBlipProcessor} for image and text preprocessing.

\subsubsection{Contrastive Decoding Configurations}
Unless otherwise specified, all contrastive decoding methods use contrast strength $\alpha=1$. All models and methods use the same generation parameters: temperature $=1$, top-$p=1$, top-$k=0$, repetition penalty $=1$, and maximum new tokens $=16$. Sampling is enabled with \texttt{do\_sample=true}.

\paragraph{VCD.}

VCD constructs its reference branch by adding diffusion noise to the preprocessed image tensor. We use a diffusion timestep of $t=500$. The original branch receives the clean image, while the reference branch receives a noised image with the same spatial dimensions and normalization.

\paragraph{ICD.}

ICD constructs the reference branch using the following disturbed instruction:

\vspace{-3pt}
\begin{quote}
\begingroup
\fontsize{8}{9.5}\selectfont
\ttfamily
You are a confused objects detector to provide a fuzzy overview or impression of the image.
\endgroup
\end{quote}
\vspace{-3pt}

For LLaVA, this instruction replaces the system prompt in the reference branch, while the user question, image, and candidates remain unchanged. For InstructBLIP, it is applied only to the Q-Former instruction of the reference branch; the prompt received by the language model remains unchanged.

\paragraph{SID.}

For LLaVA, SID ranks the 576 visual tokens using the last-query attention from the second language-model layer and retains the 100 tokens with the lowest attention in subsequent layers. For InstructBLIP, we compute visual attention at the fifth layer and retain the lowest-attention $10\%$ of the Q-Former tokens, corresponding to 3 of the 32 visual tokens. The resulting mask is applied from the sixth layer onward.

\paragraph{AvisC.}

AvisC identifies visually blind tokens using attention from the original branch. We set the layer-selection parameter to $\gamma=0.5$ and the attention threshold parameter to $\lambda=1.0$. The reference branch retains the high-attention visual tokens selected by this criterion and zeros the remaining visual representations. This operation is applied to the visual-token region for LLaVA and to the projected Q-Former visual representations for InstructBLIP.

\subsubsection{Contrast-Strength Experiment}

The default contrast strength is $\alpha=1$. For the contrast-strength experiment, we increase it to $\alpha=3$ only for LLaVA-1.5-7B under the single-ground-truth setting. All other settings, including the dataset, candidate permutations, reference branches, random seeds, and decoding parameters, remain unchanged. CGM is determined by the difference between the original and reference branches and therefore does not depend on $\alpha$. Increasing $\alpha$ only amplifies the effect of this signal on the final candidate margins and predictions.

\subsection{Effects of Adaptive Plausibility Constraint (APC) on AMBER and POPE}
\label{appendix: amber and pope}

We investigate the effect of the Adaptive Plausibility Constraint (APC) on both the AMBER generative benchmark and the POPE discriminative benchmark. Using LLaVA-1.5-7B as the base vision-language model, we evaluate four contrastive decoding methods, VCD, ICD, SID, and AvisC under controlled decoding conditions, with standard LLaVA-1.5-7B decoding serving as Baseline.

APC restricts contrastive decoding to tokens that are sufficiently probable under the original branch. Specifically, a candidate token $w$ is retained if 
\begin{equation}
    \ell_o(w) > \max_{v}\ell_o(v) + \log\beta,
\end{equation}
where $\beta$ controls the strength of the constraint. A larger $\beta$ imposes more aggressive candidate filtering. We compare $\beta=0.1$ with $\beta=0$ which denotes the unconstrained setting where APC is disabled.

\textbf{Implementation details.}
For AMBER, all methods use the same model weights, data, prompt template, random seeds, and sampling parameters.
We set the maximum context length to 2,048 tokens.

For POPE, we evaluate the model on the COCO-based Random, Popular, and Adversarial subsets. Each subset contains 3,000 questions, resulting in a total of 9,000 evaluation instances with 4,500 \emph{Yes} and 4,500 \emph{No} ground-truth answers. All methods use identical questions, images, and tokenized prompts.

\textbf{Decoding configurations.}
Both benchmarks use multinomial sampling, with different decoding configurations to accommodate their respective tasks: AMBER evaluates open-ended generation, whereas POPE requires short answers. The complete configurations are summarized in Table~\ref{tab:amber_pope_sampling_configuration}.

\begin{table}[t]
\centering
\caption{Sampling and decoding configurations used for AMBER and POPE.}
\label{tab:amber_pope_sampling_configuration}
\small
\setlength{\tabcolsep}{5pt}
\renewcommand{\arraystretch}{1.08}
\begin{tabular}{lcc}
\toprule
Parameter & AMBER & POPE \\
\midrule
Temperature & 1.0 & 1.0 \\
Top-$p$ & 0.9 & 1.0 \\
Top-$k$ & 50 & Disabled \\
Repetition penalty & 1.0 & -- \\
Contrastive strength $\alpha$ & 1.0 & 1.0 \\
APC parameter $\beta$ & 0.1 or 0 & 0.1 or 0 \\
Batch size & -- & 1 \\
Maximum generation length & 256 tokens & 8 tokens \\
VCD diffusion noise step & 500 & 500 \\
Base random seed & 42 & 42 \\
\bottomrule
\end{tabular}
\end{table}

For AMBER, temperature, top-$p$, and top-$k$ processing are applied after contrastive decoding and APC filtering. The next token is then selected through multinomial sampling.

For POPE, generation terminates early when an end-of-sequence token is produced. Generation is not explicitly restricted to a binary \emph{Yes}/\emph{No} vocabulary.

\textbf{POPE evaluation metrics.}
Generated responses are converted into \emph{Yes}/\emph{No} predictions using the same normalization and parsing procedure across methods and APC settings. 
We pool predictions across the Random, Popular, and Adversarial subsets, and report Accuracy, \emph{Yes} Accuracy (TPR, True Positive Rate), \emph{No} accuracy, FPR (False Positive Rate), \emph{Yes} Rate, and F1.

\subsection{Effects of Non-target Content on Generative Evaluation}
\label{appendix: Non-target Descriptions on Generative}
\subsubsection{Descriptive-Word Density}

We analyze free-form responses generated by LLaVA-1.5-7B for 500 MS-COCO validation images using the prompt \emph{``Please describe this image in detail.''}. We refer to this standard generation as the Baseline. We compare it with VCD, ICD, SID, AvisC, and FuzzyCD. All methods use multinomial sampling with a temperature of 1.0, top-$p=1.0$, top-$k=0$, and a maximum generation length of 256 tokens. For each image, all methods use the same deterministic sampling seed.

Responses are tokenized using NLTK's Treebank tokenizer and tagged with its English averaged-perceptron POS tagger. Punctuation is excluded from the word count. 
We define descriptive words as adjectives (\texttt{JJ}, \texttt{JJR}, and \texttt{JJS}), adverbs (\texttt{RB}, \texttt{RBR}, and \texttt{RBS}), cardinal numbers (\texttt{CD}), and spatial-relation expressions. 
The spatial lexicon includes multiword expressions, such as \emph{in front of}, \emph{next to}, and \emph{in the background}, and single-word relations, such as \emph{above}, \emph{below}, \emph{behind}, and \emph{between}. Multiword expressions are matched first to prevent double counting.

For a response $y$, descriptive-word density is defined as

\[
q(y)
=
100\,
\frac{N_{\mathrm{desc}}(y)}
     {N_{\mathrm{word}}(y)}.
\]

For method $m$, we report its mean paired difference from the Baseline:

\[
\Delta_m
=
\frac{1}{500}
\sum_{i=1}^{500}
\left[
q\!\left(y_i^m\right)
-
q\!\left(y_i^{\mathrm{base}}\right)
\right].
\]

A positive value indicates an increase in descriptive words per 100 generated words relative to the Baseline. 

\subsubsection{Controlled Prefix Intervention}

We further examine whether non-object descriptive content in the generation prefix affects hallucination in the subsequent generation. From the same 500-image Baseline run, we retain examples for which the first generated sentence contains no CHAIR object hallucination. One hundred and twenty images are selected.

For each image, we construct a neutral prefix and a descriptive prefix. Both begin with the same Baseline first sentence, followed by the bridge phrase \emph{``In addition,''}. The neutral prefix contains the following text:

\begin{quote}
\itshape
Overall, the image presents the scene in a way that remains consistent with the information already described in the preceding sentence. It continues in an ordinary manner, using familiar wording without adding any unusual claims, specific details, or new visual information.
\end{quote}

The descriptive prefix instead contains additional adjectives, color terms, and other descriptive expressions, without introducing new object mentions:

\begin{quote}
\itshape
Overall, the scene looks bright, colorful, clear, vivid, and richly detailed, with red, blue, green, and yellow tones visible throughout. The wording remains simple and direct without adding any other information here.
\end{quote}

Each constructed prefix is teacher-forced into the Baseline decoder, with the model state recomputed from the beginning before generating the continuation. The two conditions use the same image and continuation sampling seed, with a temperature of 1.0, top-$p=1.0$, top-$k=0$, and a maximum total generation length of 256 tokens. Only the generated continuation is evaluated for hallucination.

To control for differences in continuation length, continuations for each image are truncated to the shorter token length. Using the standard CHAIR object vocabulary and MS-COCO ground truth, corpus-level $\mathrm{CHAIR}_i$ is computed as $\mathrm{CHAIR}_i=\sum_i H_i/\sum_i O_i$, where $H_i$ and $O_i$ denote the numbers of hallucinated and total object mentions, respectively. $\mathrm{CHAIR}_s$ is computed as the
proportion of images containing at least one hallucinated object in the continuations.

\paragraph{Results.}
The neutral-prefix condition produces 32 hallucinated mentions among 135 object mentions, with hallucinations occurring in 24 of 120 continuations, yielding $\mathrm{CHAIR}_i=23.70\%$ and $\mathrm{CHAIR}_s=20.00\%$. 
In comparison, the descriptive-prefix condition produces 47 hallucinated mentions among 195 object mentions and hallucinations in 38 of 120 continuations, yielding $\mathrm{CHAIR}_i=24.10\%$ and $\mathrm{CHAIR}_s=31.67\%$.

\subsection{Effects of Method-Generated Prefixes on Hallucinations in Subsequent Generation}

\begin{table}[t]
\centering
\caption{Full-response hallucination evaluation results of LLaVA-1.5-7B on AMBER generative tasks.}
\label{tab:amber_generation}
\small
\begin{tabular*}{\columnwidth}{@{\extracolsep{\fill}}lcccc@{}}
\toprule
Method
& CH$\downarrow$
& Cover$\uparrow$
& Hal$\downarrow$
& Cog$\downarrow$ \\
\midrule
Baseline & 10.6 & 50.7 & 46.4 & 4.7 \\
VCD      & 10.5 & 51.5 & 46.4 & 4.5 \\
ICD      & 8.9  & 51.5 & 40.2 & 4.0 \\
SID      & \textbf{8.0} & 52.2 & \textbf{39.0} & \textbf{3.3} \\
AvisC    & 10.7 & \textbf{52.3} & 48.0 & 4.8 \\
\bottomrule
\end{tabular*}
\end{table}

Table~\ref{tab:amber_generation} shows clear differences among dehallucination methods in full-response evaluation. SID achieves the best overall performance, with CH, Hal, and Cog scores of 8.0, 39.0, and 3.3, respectively. ICD also outperforms the baseline, obtaining corresponding scores of 8.9, 40.2, and 4.0, while VCD performs comparably to the baseline. In contrast, although AvisC achieves the highest Cover score of 52.3, its CH, Hal, and Cog scores are 10.7, 48.0, and 4.8, respectively, all slightly worse than the baseline scores of 10.6, 46.4, and 4.7. To explain the relatively weak full-response performance of AvisC, Figure~\ref{fig:prefix} further examines how method-generated prefixes affect the hallucination rate of subsequent generation.

When the baseline continuation is held fixed, replacing the base prefix with the corresponding method prefix increases the hallucination rate for all four methods. Specifically, the hallucination rate increases from 21.26\% to 22.51\% for VCD, an increase of 1.25 percentage points; from 21.26\% to 22.45\% for ICD, an increase of 1.19 percentage points; and from 21.26\% to 22.50\% for SID, an increase of 1.24 percentage points. For AvisC, the hallucination rate increases from 21.26\% to 23.28\%, corresponding to an increase of 2.02 percentage points. These results indicate that method-generated prefixes generally introduce interference into subsequent decoding, thereby increasing the hallucination rate of the continuation. Notably, the increase caused by the AvisC prefix is substantially larger than those caused by the other methods.

The same trend persists when each method's own continuation strategy is held fixed. Relative to the base prefix, the method-generated prefixes of VCD, ICD, SID, and AvisC increase the hallucination rate by 0.85, 0.61, 0.19, and 1.44 percentage points, respectively. The SID prefix has the smallest effect, whereas the AvisC prefix again produces the largest increase in hallucination rate. The consistent results across both comparisons demonstrate that prefix replacement itself is an important factor contributing to increased hallucinations in subsequent generation, although the magnitude of this effect varies across methods.

\begin{figure}[H]
\centering
\includegraphics[width=\columnwidth]{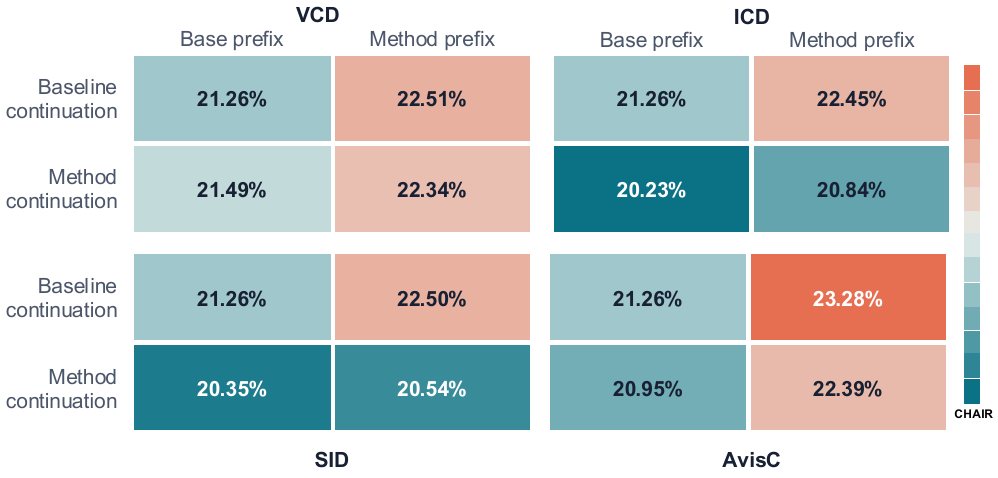}
\caption{Prefix-continuation factorial experiment. Entries are the reported hallucination rate. ``Base'' and ``Method'' denote which continuation is decoded after the fixed prefix.}
\label{fig:prefix}
\end{figure}

\subsection{Robustness Across Different Candidate-Selection Settings}
\label{appendix: across options settings}
The construction of candidate options can change the difficulty of competition between ground-truth and hallucinated candidates. We therefore examine whether the observed differences among methods are robust to different candidate-selection strategy. Table~\ref{tab:cgm_option_settings} compares four settings: scene-related, co-occurrence, random, and balanced. The baseline accuracy varies from 58.2\% to 70.6\% across these settings. In particular, the co-occurrence setting yields the lowest baseline accuracy, whereas the random setting yields the highest, indicating that the different candidate-construction strategies produce competition of  varying difficulty.

Despite these variations in task difficulty, the direction and ordering of CGM remain \textbf{consistent} across all four settings. VCD consistently produces negative CGM values of $-0.106$, $-0.068$, $-0.034$, and $-0.084$ under the scene-related, co-occurrence, random, and balanced settings, respectively, with Positive rates ranging only from 36.8\% to 40.8\%. In contrast, ICD, SID, and AvisC achieve positive CGM values under every setting. ICD obtains CGM values ranging from 0.368 to 0.421, with Positive rates of 91.1\%--95.0\%; SID obtains CGM values of 1.015--1.102, with Positive rates of 66.8\%--73.0\%; and AvisC achieves the largest CGM values, ranging from 3.005 to 3.844, together with consistently high Positive rates of 97.6\%--99.4\%. Consequently, the CGM ordering remains $\text{AvisC} > \text{SID} > \text{ICD} > \text{VCD}$ across all four settings, demonstrating that this ordering does not depend on any particular candidate-selection strategies.

\begin{table}[t]
\centering
\caption{Robustness across different candidate-selection settings. Mean CGM, positive-CGM rate, and accuracy are reported under four candidate-selection settings.}
\label{tab:cgm_option_settings}

\small
\renewcommand{\arraystretch}{1.05}

\begin{tabular*}{\columnwidth}{
    @{\extracolsep{\fill}}llccc@{}
}
\toprule
Option Setting
& Method
& CGM
& Positive-CGM
& Acc. \\
\midrule

\multirow{5}{*}{Scene-related}
& Baseline & \textemdash & \textemdash & 59.1\% \\
& VCD   & $-0.106$ & 39.8\% & 58.0\% \\
& ICD   & $0.420$  & 93.3\% & 64.4\% \\
& SID   & $1.015$  & 70.6\% & 68.7\% \\
& AvisC & $3.005$  & 97.6\% & 73.6\% \\
\midrule

\multirow{5}{*}{Co-occurrence}
& Baseline & \textemdash & \textemdash & 58.2\% \\
& VCD   & $-0.068$ & 40.8\% & 56.5\% \\
& ICD   & $0.421$  & 92.9\% & 63.0\% \\
& SID   & $1.088$  & 73.0\% & 69.6\% \\
& AvisC & $3.040$  & 97.6\% & 73.0\% \\
\midrule

\multirow{5}{*}{Random}
& Baseline & \textemdash & \textemdash & 70.6\% \\
& VCD   & $-0.034$ & 36.8\% & 69.9\% \\
& ICD   & $0.368$  & 95.0\% & 75.1\% \\
& SID   & $1.078$  & 66.8\% & 78.7\% \\
& AvisC & $3.844$  & 99.4\% & 86.1\% \\
\midrule

\multirow{5}{*}{Balanced}
& Baseline & \textemdash & \textemdash & 65.6\% \\
& VCD   & $-0.084$ & 37.1\% & 63.6\% \\
& ICD   & $0.382$  & 91.1\% & 69.9\% \\
& SID   & $1.102$  & 71.4\% & 73.0\% \\
& AvisC & $3.354$  & 98.1\% & 79.2\% \\
\bottomrule
\end{tabular*}
\end{table}

\section{Robustness to Annotation Missingness}
\label{sec:coco-missingness-robustness}

COCO object annotations are not exhaustive. Consequently, an object that is present in an image but unannotated may be incorrectly treated as absent.
This issue can affect not only the estimated object co-occurrence statistics but also the resulting hallucinated candidates. 
We therefore examine whether such annotation missingness alters our conclusions about contrastive correction produced by different methods.

\paragraph{Experimental setup.}
We conduct the experiment on a newly sampled set of 200 images and simulate incomplete annotations by artificially removing object-instance annotations. We consider two missingness mechanisms. Under \emph{uniform missingness}, every eligible object instance has the same probability of being removed. Under \emph{small-object-prioritized missingness}, smaller instances are more likely to be removed, approximating the greater likelihood that small objects are missed in image annotations. We consider missingness rates of $q=10\%$ and $q=20\%$, where $q$ denotes the proportion of eligible object-instance annotations removed.

Because different sets of removed annotations can affect benchmark construction differently even at the same missingness rate, we further examine this variation for $q=10\%$.
For each missingness mechanism, we generate 50 independent \emph{missingness realizations}, each corresponding to a different set of annotations removed at the same $q=10\%$ rate.
We characterize the impact of each realization using two criteria.
First, \emph{contaminated questions} are questions in which an object removed from the annotations, but actually present in the image, is subsequently selected as a hallucinated candidate.
Second, \emph{changed questions} are questions in which at least one structured hallucinated candidate differs from the $q=0$ setting.
We rank the realizations first by the number of contaminated questions and then by the number of changed questions, and select the P10, P50, and P90 realizations as the \emph{mild}, \emph{median}, and \emph{severe} conditions, respectively. 
These conditions have the same missingness rate but differ in how strongly the particular pattern of missing annotation affects benchmark construction.

For every condition, we re-derive the image-level object sets, recompute the category co-occurrence probabilities, and reconstruct the three structured hallucinated candidates. The random hallucinated candidate is held fixed as a control. Each question is evaluated with five cyclic answer-position permutations. We evaluate Baseline, VCD, ICD, SID, and AvisC using LLaVA-1.5-7B with $\alpha=1$ and $\beta=0$ for all CD method.

\begin{table}[t]
\centering
\caption{Robustness under uniform and small-object-prioritized instance missingness.}
\label{tab:missingness-robustness}
\vspace{2pt}
\scriptsize
\setlength{\tabcolsep}{2.5pt}
\renewcommand{\arraystretch}{1.03}
\resizebox{\columnwidth}{!}{%
\begin{tabular}{@{}llccc@{\hspace{9pt}}ccc@{}}
\toprule
& & \multicolumn{3}{c}{Uniform} &
\multicolumn{3}{c}{Small-object-prioritized} \\
\cmidrule(l{3pt}r{7pt}){3-5}
\cmidrule{6-8}
Condition & Method & CGM & Positive-CGM & Acc. & CGM & Positive-CGM & Acc. \\
\midrule
\multirow{5}{*}{$q=0$}
 & Baseline & -- & -- & 70.0 & -- & -- & 70.0 \\
 & VCD   & 0.031 & 44.1 & 68.5 & 0.031 & 44.1 & 68.5 \\
 & ICD   & 0.377 & 93.9 & 72.0 & 0.377 & 93.9 & 72.0 \\
 & SID   & 1.250 & 72.5 & 76.5 & 1.250 & 72.5 & 76.5 \\
 & AvisC & 3.251 & 98.4 & 80.5 & 3.251 & 98.4 & 80.5 \\
\midrule

\multirow{5}{*}{\shortstack{$q=10\%$\\mild}}
 & Baseline & -- & -- & 71.5 & -- & -- & 68.5 \\
 & VCD   & 0.030 & 43.8 & 70.5 & 0.028 & 44.3 & 69.0 \\
 & ICD   & 0.384 & 94.5 & 77.5 & 0.378 & 94.5 & 72.5 \\
 & SID   & 1.282 & 72.6 & 82.0 & 1.267 & 72.1 & 78.5 \\
 & AvisC & 3.251 & 98.9 & 82.0 & 3.246 & 98.5 & 81.5 \\
\midrule

\multirow{5}{*}{\shortstack{$q=10\%$\\median}}
 & Baseline & -- & -- & 68.0 & -- & -- & 71.0 \\
 & VCD   & 0.031 & 43.8 & 67.0 & 0.023 & 44.1 & 72.0 \\
 & ICD   & 0.382 & 93.5 & 72.0 & 0.374 & 94.4 & 74.0 \\
 & SID   & 1.270 & 73.9 & 79.5 & 1.286 & 72.5 & 80.0 \\
 & AvisC & 3.258 & 98.6 & 78.5 & 3.243 & 98.5 & 81.0 \\
\midrule

\multirow{5}{*}{\shortstack{$q=10\%$\\severe}}
 & Baseline & -- & -- & 65.0 & -- & -- & 64.0 \\
 & VCD   & 0.040 & 44.4 & 64.5 & 0.032 & 44.4 & 66.0 \\
 & ICD   & 0.381 & 93.4 & 74.0 & 0.374 & 93.6 & 73.0 \\
 & SID   & 1.275 & 73.0 & 78.0 & 1.280 & 72.8 & 76.5 \\
 & AvisC & 3.243 & 98.5 & 80.0 & 3.244 & 98.3 & 80.0 \\
\midrule

\multirow{5}{*}{\shortstack{$q=20\%$\\median}}
 & Baseline & -- & -- & 71.0 & -- & -- & 69.5 \\
 & VCD   & 0.044 & 45.0 & 67.0 & 0.035 & 46.3 & 69.0 \\
 & ICD   & 0.372 & 93.9 & 74.5 & 0.380 & 93.9 & 75.0 \\
 & SID   & 1.214 & 72.3 & 77.5 & 1.315 & 73.8 & 78.5 \\
 & AvisC & 3.229 & 98.3 & 80.0 & 3.239 & 98.4 & 80.5 \\
\bottomrule
\end{tabular}
}
\end{table}

\paragraph{Results.}
Table~\ref{tab:missingness-robustness} reports the results under uniform and small-object-prioritized missingness. Across both missingness mechanisms, the main CGM patterns remain highly stable.

AvisC consistently achieves the largest CGM, with mean values ranging from 3.229 to 3.258 and positive-CGM ranging from 98.3\% to 98.9\%. The CGM of SID remains between 1.214 and 1.315, clearly exceeding that of ICD, while its positive-CGM ranges from 72.1\% to 73.9\%. Although ICD has a smaller CGM, its positive-CGM remains between 93.4\% and 94.5\%, indicating that its reference branch provides a highly consistent corrective direction. In contrast, the CGM of VCD remains close to zero and its positive-CGM is consistently below 50\%, suggesting that its reference branch does not provide a consistently beneficial contrastive direction.

More importantly, the ordering by mean CGM remains unchanged across all nine conditions:
\[
\mathrm{AvisC} > \mathrm{SID} > \mathrm{ICD} > \mathrm{VCD}.
\]
Thus, whether missing annotations are simulated uniformly or prioritized toward small objects to better approximate realistic annotation bias, recomputing the co-occurrence statistics and reconstructing the distractors does \emph{not} alter the CGM-based ranking of the CD methods. The positive-CGM ordering is also stable across all conditions, following 
\[
\mathrm{AvisC} > \mathrm{ICD} > \mathrm{SID} > \mathrm{VCD}.
\]

\begin{table}[t]
\centering
\caption{Robustness under three datasets.}
\label{tab:three_dataset_cgm}
\small
\setlength{\tabcolsep}{2.5pt}
\renewcommand{\arraystretch}{1.05}
\begin{tabularx}{\columnwidth}{
    @{}
    l
    l
    >{\centering\arraybackslash}X
    >{\centering\arraybackslash}X
    >{\centering\arraybackslash}X
    @{}
}
\toprule
Dataset & Method & CGM & Positive-CGM & Acc. \\
\midrule

\multirow{5}{*}{COCO}
& Baseline & \textemdash & \textemdash & 65.6\% \\
& VCD      & -0.084 & 37.1\% & 63.6\% \\
& ICD      &  0.382 & 91.1\% & 69.9\% \\
& SID      &  1.102 & 71.4\% & 73.0\% \\
& AvisC    &  3.354 & 98.1\% & 79.2\% \\

\midrule

\multirow{5}{*}{Open Images}
& Baseline & \textemdash & \textemdash & 68.4\% \\
& VCD      & -0.055 & 42.6\% & 67.3\% \\
& ICD      &  0.424 & 94.7\% & 73.8\% \\
& SID      &  1.431 & 76.6\% & 78.1\% \\
& AvisC    &  3.439 & 98.8\% & 83.5\% \\

\midrule

\multirow{5}{*}{Objects365}
& Baseline & \textemdash & \textemdash & 54.6\% \\
& VCD      & -0.114 & 40.0\% & 53.7\% \\
& ICD      &  0.345 & 90.1\% & 59.7\% \\
& SID      &  1.096 & 69.8\% & 68.4\% \\
& AvisC    &  2.753 & 94.1\% & 69.2\% \\

\bottomrule
\end{tabularx}
\end{table}

\begin{table*}[t]
\centering
\small
\caption{
Conditional co-occurrence scores used to construct the hallucinated candidates in Figure~\ref{fig:candidate_example}. 
Each entry reports $C(a,o)=N(a,o)/N(a)$ between an annotated category $a$ and an unannotated candidate $o$. The final rows report the resulting $S_{\mathrm{scene}}$ and $S_{\max}$ scores and the candidate type assigned to each selected hallucinated candidate.
}
\label{tab:cooccurrence_example}
\begin{tabular}{lccccc}
\toprule
Annotated category $a$ &
$N(a)$ &
\textit{person} &
\textit{dining table} &
\textit{zebra} &
\textit{sink} \\
\midrule
\textit{backpack}
    & 1,832 & 0.9039 & 0.0535 & 0.0011 & 0.0076 \\
\textit{book}
    & 1,828 & 0.4152 & 0.2030 & 0.0000 & 0.0481 \\
\textit{cell phone}
    & 1,695 & 0.7900 & 0.1504 & 0.0006 & 0.0177 \\
\textit{knife}
    & 1,410 & 0.5312 & 0.6170 & 0.0007 & 0.1121 \\
\midrule
$S_{\mathrm{scene}}$
    & -- & \textbf{0.6601} & \textbf{0.2560}
    & \textbf{0.0006} & \textbf{0.0464} \\
$S_{\mathrm{max}}$
    & -- & \textbf{0.9039} & \textbf{0.6170}
    & \textbf{0.0011} & \textbf{0.1121} \\
Candidate type
    & -- & Highly co-occurring & Scene-related
    & Scene-unrelated & Random \\
\bottomrule
\end{tabular}
\end{table*}

\subsection{Robustness Across Datasets}
\label{appendix: across datasets}

Because \concor{} is instantiated on COCO, we examine whether the observed differences among CD methods are specific to its image and object distribution. We repeat the evaluation on Open Images~\cite{kuznetsova2020open} and Objects365~\cite{shao2019objects365}, which provide different image sources and object distributions, to assess whether the main CGM patterns generalize across datasets.

Table~\ref{tab:three_dataset_cgm} summarizes the results of LLaVA-1.5-7B on 200 images from each of the COCO, Open Images, and Objects365. All three experiments follow the same five-choice, single-ground-truth MCQ setting and compare Baseline with four CD methods: VCD, ICD, SID, and AvisC.

The relative patterns among methods are consistent across all three datasets.  In terms of CGM, the four CD methods follow the same ordering:
$\mathrm{AvisC} > \mathrm{SID} > \mathrm{ICD} > \mathrm{VCD}$.
AvisC achieves the largest mean CGM on every dataset, followed by SID and ICD, whereas VCD produces negative mean CGM across all three datasets.

The ordering by positive-CGM rate is also consistent:
$\mathrm{AvisC} > \mathrm{ICD} > \mathrm{SID} > \mathrm{VCD}$.
AvisC achieves positive-CGM rates above 94\% on all three datasets, while ICD remains above 90\%. Although its mean CGM is lower than that of SID, its higher positive-CGM rate  indicates more consistent improvements across samples. SID produces a larger average gain but exhibits greater sample-level variation.

The five-candidate sampling accuracy follows the same ordering on all three datasets:
$\mathrm{AvisC} > \mathrm{SID} > \mathrm{ICD} > \mathrm{Baseline} > \mathrm{VCD}$.
AvisC achieves the highest accuracy on every dataset, while SID and ICD consistently outperform Baseline. In contrast, VCD performs slightly below Baseline in all three datasets. 
Overall, the consistent method orderings across datasets show that the main differences identified by CGM are robust to the tested image sources rather than being specific to COCO.

\section{Example of Hallucinated Candidate Construction}
\label{appendix: didate Construction Example}
Figure~\ref{fig:candidate_example} presents an example from \concor{} (COCO image ID 37900). The designated ground-truth object is
\textit{backpack}.
The set of COCO-annotated object categories in this image is

\begin{equation}
P_I =
\{
\textit{backpack},
\textit{book},
\textit{cell phone},
\textit{knife}
\}.
\end{equation}

For each unannotated candidate category $o$, we compute its conditional co-occurrence with each annotated category $a\in P_I$ as
$C(a,o)=N(a,o)/N(a)$ where $N(a)$ is the number of COCO images containing category $a$, and $N(a,o)$ is the number containing both $a$ and $o$. 
Table~\ref{tab:cooccurrence_example} reports these values for the four selected hallucinated candidates.

\begin{figure}[t]
    \centering
    \includegraphics[width=\linewidth]{figures/candidate_construction_example.jpg}
    \caption{
    Candidate-construction example from COCO Val2014.
    The designated ground-truth candidate is \textit{backpack}, while
    \textit{person}, \textit{dining table}, \textit{zebra}, and
    \textit{sink} are selected as the highly co-occurring, scene-related, scene-unrelated, and random hallucinated candidates, respectively.
    }
    \label{fig:candidate_example}
\end{figure}

The candidate \textit{person} has the highest $S_{\mathrm{max}}$, $0.9039$, and is therefore selected as the highly co-occurring candidate. After excluding \textit{person}, \textit{dining table} has the highest $S_{\mathrm{scene}}$, $0.2560$, and is selected as the scene-related candidate. 
The co-occurrence values of \textit{zebra} are close to zero for every annotated category, making it the scene-unrelated candidate. Finally, \textit{sink} is uniformly sampled from the remaining unannotated categories as the random candidate.

\end{document}